\documentclass{article} % For LaTeX2e
\usepackage{iclr2017_conference,times}
\pdfoutput=1
\usepackage{hyperref}
\usepackage{url}

\usepackage[utf8]{inputenc} % allow utf-8 input
\usepackage[T1]{fontenc}    % use 8-bit T1 fonts
\usepackage{hyperref}       % hyperlinks
\usepackage{url}            % simple URL typesetting
\usepackage{booktabs}       % professional-quality tables
\usepackage{amsfonts}       % blackboard math symbols
\usepackage{nicefrac}       % compact symbols for 1/2, etc.
\usepackage{microtype}      % microtypography
\usepackage{graphicx}
\usepackage{subfigure}
\usepackage{caption}
\usepackage{amsmath}

\usepackage[lined,boxed,ruled, commentsnumbered]{algorithm2e}
\usepackage{dsfont}
\usepackage{comment}
\usepackage{tabularx}
\usepackage{color}
\usepackage{footnote}
\usepackage{multirow}

\title{Trained Ternary Quantization}

\author{Chenzhuo Zhu\thanks{Work done while at Stanford CVA lab.} \\
Tsinghua University \\
\texttt{zhucz13@mails.tsinghua.edu.cn} \\
\And
\hspace{-200pt}
Song Han \\
\hspace{-200pt}
Stanford University \\
\hspace{-200pt}
\texttt{songhan@stanford.edu} \\
\AND
Huizi Mao \\
Stanford University \\
\texttt{huizi@stanford.edu\hspace{3pt}} \\
\And
William J. Dally \\
Stanford University \\
NVIDIA \\
\texttt{dally@stanford.edu}
}

\begin{document}

\iclrfinalcopy
\maketitle

\begin{abstract}
Deep neural networks are widely used in machine learning applications. However, the deployment of large neural networks models can be difficult to deploy on mobile devices with limited  power budgets. To solve this problem, we propose Trained Ternary Quantization~(TTQ), a method that can reduce the precision of weights in neural networks to ternary values. This method has very little accuracy degradation and can even improve the accuracy of some models (32, 44, 56-layer ResNet) on CIFAR-10 and AlexNet on ImageNet. And our AlexNet model is trained from scratch, which means it's as easy as to train normal full precision model. We highlight our trained quantization method that can learn both ternary values and ternary assignment. During inference, only ternary values (2-bit weights) and scaling factors are needed, therefore our models are nearly $16\times $ smaller than full-precision models. Our ternary models can also be viewed as sparse binary weight networks, which can potentially be accelerated with custom circuit. Experiments on CIFAR-10 show that the ternary models obtained by trained quantization method \emph{outperform} full-precision models of ResNet-32,44,56 by 0.04\%, 0.16\%, 0.36\%, respectively. On ImageNet, our model outperforms full-precision AlexNet model by 0.3\% of Top-1 accuracy and outperforms previous ternary models by 3\%.

% we propose a quantization method that can lower the precision of weights in neural networks to ternary values, which can efficiently reduce energy consumption from both computation and memory access. By trained quantization of positive and negative ternary weight, we are able to recover the loss of accuracy brought by reduced precision. Experiments on CIFAR-10 show that the ternary model obtained from our trained quantization method outperforms full precision ResNet model by 0.1\%$\sim$0.3\%.
\end{abstract}
% In this paper, we propose a novel deep neural network quantization method to lower the weight precision. We train two additional variables for each layer used as quantized values together with other parameters. During inference, only ternary value weights and layer-wise quantized values are needed therefore our model can be $16\times $ smaller than full precision model. Our tenary models can also be viewed as sparse binary weight networks, which can be speed up through special designed circuits therefore making it possible to deploy very deep neural networks on small devices. Furthermore, training with adaptive quantization values greatly enhances capacity of quantized ternary-weight models. Our experiments on CIFAR-10 show that the ternary model obtained from trained quantization method outperforms full precision ResNet model by 0.1\%$\sim$0.3\%.
\section{Introduction}

Deep neural networks are becoming the preferred approach for many machine learning applications. However, as networks get deeper, deploying a network with a large number of parameters on a small device becomes increasingly difficult. Much work has been
done to reduce the size of networks.  Half-precision networks~\citep{amodei2015deep} cut sizes of neural networks in half. XNOR-Net~\citep{rastegari2016xnor}, DoReFa-Net~\citep{zhou2016dorefa} and network binarization~\citep{courbariauxbinarized,courbariaux2015binaryconnect,lin2015neural} use aggressively quantized weights, activations and gradients to further reduce computation during training. While weight binarization benefits from $32\times$ smaller model size, the extreme compression rate comes with a loss of accuracy. \cite{hubara2016quantized} and \cite{li2016ternary} propose ternary weight networks to trade off between model size and accuracy. 

In this paper, we propose Trained Ternary Quantization which uses two full-precision scaling coefficients $W^p_l$, $W^n_l$ for each layer $l$, and quantize the weights to \{$-W^n_l$, 0, $+W^p_l$\} instead of traditional \{-1, 0, +1\} or \{-E, 0, +E\} where E is the mean of the absolute weight value, which is not learned. Our positive and negative weights have different absolute values $W^p_l$ and $W^n_l$ that are trainable parameters. We also maintain latent full-precision weights at training time, and discard them at test time. We back propagate the gradient to both $W^p_l$, $W^n_l$ and to the latent full-precision weights.  This makes it possible to adjust the ternary assignment (i.e. which of the three values a weight is assigned).

Our quantization method, achieves higher accuracy on the CIFAR-10 and ImageNet datasets. For AlexNet on ImageNet dataset, our method outperforms previously state-of-art ternary network\citep{li2016ternary} by 3.0\% of Top-1 accuracy and the full-precision model by 1.6\%. By converting most of the parameters to 2-bit values, we also compress the network by about 16x. Moreover, the advantage of few multiplications still remains, because $W^p_l$ and $W^n_l$ are fixed for each layer during inference. On custom hardware, multiplications can be pre-computed on activations, so only two multiplications per activation are required.

% \textcolor{red}{However, this is at the cost of bringing back multipliers. $W^i_n$ and $W^i_p$ have different values, so multiplications are necessary. We believe this trade-off is worthwhile because memory reference takes two orders of magnitude more energy than multiplication in silicon~\cite{han2016eie}}.

% However, in silicon design, multiplication takes 100$\times$ less energy compared with memory reference~\cite{han2015learning}. According to \cite{han2016eie}, the area and energy consumption of multipliers are very small compared to memory access. So, it's worthwhile to have different $W^i_n$ / $W^i_p$ and trade for accuracy.

\section{Motivations}
%As centralized, server-mount artificial intelligence systems have already benefited a number of real-world applications (AlphaGo for example),
% 前面写的不太好
% 直接从mobile deployment开始写？大网络的性能感觉之前提过无数遍了。。
The potential of deep neural networks, once deployed to mobile devices, has the advantage of lower latency, no reliance on the network, and better user privacy. However, energy efficiency becomes the bottleneck for deploying deep neural networks on mobile devices because mobile devices are battery constrained. Current deep neural network models consist of hundreds of millions of parameters. Reducing the size of a DNN model makes the deployment on edge devices easier. 

First, a smaller model means less overhead when exporting models to clients. Take autonomous driving for example; Tesla periodically copies new models from their servers to customers’ cars. Smaller models require less communication in such over-the-air updates, making frequent updates more feasible. Another example is on Apple Store; apps above 100 MB will not download until you connect to Wi-Fi. It’s infeasible to put a large DNN model in an app. The second issue is energy consumption. Deep learning is energy consuming, which is problematic for battery-constrained mobile devices. As a result, iOS 10 requires iPhone to be plugged with charger while performing photo analysis. Fetching DNN models from memory takes more than two orders of magnitude more energy than arithmetic operations. Smaller neural networks require less memory bandwidth to fetch the model, saving the energy and extending battery life. The third issue is area cost. When deploying DNNs on Application-Specific Integrated Circuits (ASICs), a sufficiently small model can be stored directly on-chip, and smaller models enable a smaller ASIC die.

% 没有提到乘法器部分，感觉在这里说我们re-introduce了乘法器不太好说。。
Several previous works aimed to improve energy and spatial efficiency of deep networks. One common strategy proven useful is to quantize 32-bit weights to one or two bits, which greatly reduces model size and saves memory reference. However, experimental results show that compressed weights usually come with degraded performance, which is a great loss for some performance-sensitive applications. The contradiction between compression and performance motivates us to work on trained ternary quantization, minimizing performance degradation of deep neural networks while saving as much energy and space as possible.

\section{Related Work}
\subsection{Binary Neural Network (BNN)}
\cite{lin2015neural} proposed binary and ternary connections to compress neural networks and speed up computation during inference. They used similar probabilistic methods to convert 32-bit weights into binary values or ternary values, defined as:
\begin{equation} \label{eq:1}
\begin{split}
w^b\sim {\rm Bernoulli}(\frac{\tilde{w}+1}{2})\times{2}-1 \\
w^t\sim {\rm Bernoulli}(|\tilde{w}|)\times{\text{sign}(\tilde{w})}
\end{split}
\end{equation}
Here $w^b$ and $w^t$ denote binary and ternary weights after quantization. $\tilde{w}$ denotes the latent full precision weight.

During back-propagation, as the above quantization equations are not differentiable, derivatives of expectations of the Bernoulli distribution are computed instead, yielding the identity function: 
\begin{equation} \label{eq:2}
\frac{\partial L}{\partial \tilde{w}}
=\frac{\partial L}{\partial w^b}
=\frac{\partial L}{\partial w^t}
\end{equation}
Here $L$ is the loss to optimize.

For BNN with binary connections, only quantized binary values are needed for inference. Therefore a $32\times$ smaller model can be deployed into applications.

\subsection{DoReFa-Net}
\cite{zhou2016dorefa} proposed DoReFa-Net which quantizes weights, activations and gradients of neural networks using different widths of bits. Therefore with specifically designed low-bit multiplication algorithm or hardware, both training and inference stages can be accelerated. 

They also introduced a much simpler method to quantize 32-bit weights to binary values, defined as:
\begin{equation} \label{eq:3}
w^b=\boldsymbol{E}(|\tilde{w}|)\times \text{sign}(\tilde{w})
\end{equation}
Here $\boldsymbol{E}(|\tilde{w}|)$ calculates the mean of absolute values of full precision weights $\tilde{w}$ as layer-wise scaling factors. During back-propagation, Equation \ref{eq:2} still applies.

\subsection{Ternary Weight Networks}
\cite{li2016ternary} proposed TWN (Ternary weight networks), which reduce accuracy loss of binary networks by introducing zero as a third quantized value. They use two symmetric thresholds $\pm \Delta_l$ and a scaling factor $W_l$ for each layer $l$ to quantize weighs into $\{-W_l, 0, +W_l\}$: 
\begin{equation} \label{eq:4}
w^t_l=\left\{
\begin{aligned}
W_l & : \tilde{w}_l>\Delta_l \\
0 & : |\tilde{w}_l|\leq \Delta_l \\
-W_l & : \tilde{w}_l<-\Delta_l 
\end{aligned}
\right.
\end{equation}
They then solve an optimization problem of minimizing L2 distance between full precision and ternary weights to obtain layer-wise values of $W_l$ and $\Delta_l$:
\begin{equation} \label{eq:5}
\begin{split}
\Delta_l &=0.7\times \boldsymbol{E}(|\tilde{w}_l|) \\
W_l &=\underset{i\in \{i|\tilde{w}_l(i)|>\Delta\}}{\boldsymbol{E}}(|\tilde{w}_l(i)|)
\end{split}
\end{equation}
And again Equation \ref{eq:2} is used to calculate gradients.
While an additional bit is required for ternary weights, TWN achieves a validation accuracy that is very close to full precision networks according to their paper.

\subsection{Deep Compression}
\cite{han2015deep} proposed deep compression to prune away trivial connections and reduce precision of weights. 
Unlike above models using zero or symmetric thresholds to quantize high precision weights, Deep Compression used clusters to categorize weights into groups. In Deep Compression, low precision weights are fine-tuned from a pre-trained full precision network, and the assignment of each weight is established at the beginning and stay unchanged, while representative value of each cluster is updated throughout fine-tuning.

\section{Method}
Our method is illustrated in Figure~\ref{fig:process}. First, we normalize the full-precision weights to the range [-1, +1] by dividing each weight by the maximum weight. Next, we quantize the intermediate full-resolution weights to \{-1, 0, +1\} by thresholding. The threshold factor $t$ is a hyper-parameter that is the same across all the layers in order to reduce the search space. Finally, we perform trained quantization by  back propagating two gradients, as shown in the dashed lines in Figure~\ref{fig:process}. We back-propagate $gradient_1$ to the full-resolution weights and $gradient_2$ to the scaling coefficients. The former enables learning the ternary \textbf{assignments}, and the latter enables learning the ternary \textbf{values}.

At inference time, we throw away the full-resolution weights and only use ternary weights.

\subsection{Learning both Ternary Values and Ternary Assignments}

\begin{figure}[t]
% \vspace{-20pt}
\hspace{-13pt}
\centering
\includegraphics[width=1.02\textwidth]{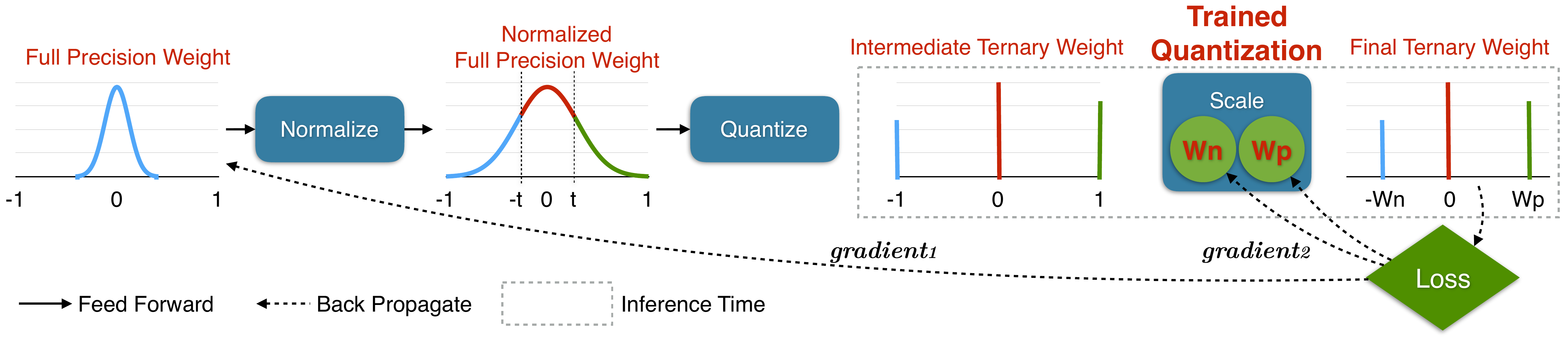}

% note: The trained quantization is similar to Deep Compression, but Deep Compression only have gradient_2, thus the cluster assignment is fixed. Here we also have gradient_1 which allows cluster assignment to change. 

\caption{Overview of the trained ternary quantization procedure.}
\label{fig:process}
% \vspace{-20pt}
\end{figure}

During gradient descent we learn both the quantized ternary weights~(the codebook), and choose which of these values is assigned to each weight (choosing the codebook index). 

To learn the ternary value~(codebook), we introduce two quantization factors $W^p_l$ and $W^n_l$ for positive and negative weights in each layer $l$. During feed-forward, quantized ternary weights $w^t_l$ are calculated as: 
\begin{equation}
w^t_l=\left\{
\begin{aligned}
W^p_l & : \tilde{w}_l>\Delta_l \\
0 & : |\tilde{w}_l|\leq \Delta_l \\
-W^n_l & : \tilde{w}_l<-\Delta_l 
\end{aligned}
\right.
\end{equation}
\label{eqn:weight-assign}

Unlike previous work where quantized weights are calculated from 32-bit weights, the scaling coefficients $W^p_l$ and $W^n_l$ are two independent parameters and are trained together with other parameters. Following the rule of gradient descent, derivatives of $W^p_l$ and $W^n_l$ are calculated as: 
\begin{equation}
\frac{\partial L}{\partial W^p_l} = \sum_{i\in I^p_l} \frac{\partial L}{\partial w^t_l(i)},
\frac{\partial L}{\partial W^n_l} = \sum_{i\in I^n_l} \frac{\partial L}{\partial w^t_l(i)}
\end{equation}
Here $I^p_l=\{i|\tilde{w}_l(i)>\Delta_l\}$ and $I^n_l=\{i|(i)\tilde{w}_l<-\Delta_l\}$. Furthermore, because of the existence of two scaling factors, gradients of latent full precision weights can no longer be calculated by Equation~\ref{eq:2}. We use scaled gradients for 32-bit weights:
\begin{equation}
\frac{\partial L}{\partial \tilde{w}_l}=\left\{
\begin{aligned}
W^p_l\times \frac{\partial L}{\partial w^t_l} & : \tilde{w}_l>\Delta_l \\
1\times \frac{\partial L}{\partial w^t_l} & : |\tilde{w}_l|\leq \Delta_l \\
W^n_l\times \frac{\partial L}{\partial w^t_l} & : \tilde{w}_l<-\Delta_l 
\end{aligned}
\right.
\end{equation}

\begin{figure}[t]
% \vspace{-10pt}
\hspace{-13pt}
\centering
\includegraphics[width=0.9\textwidth]{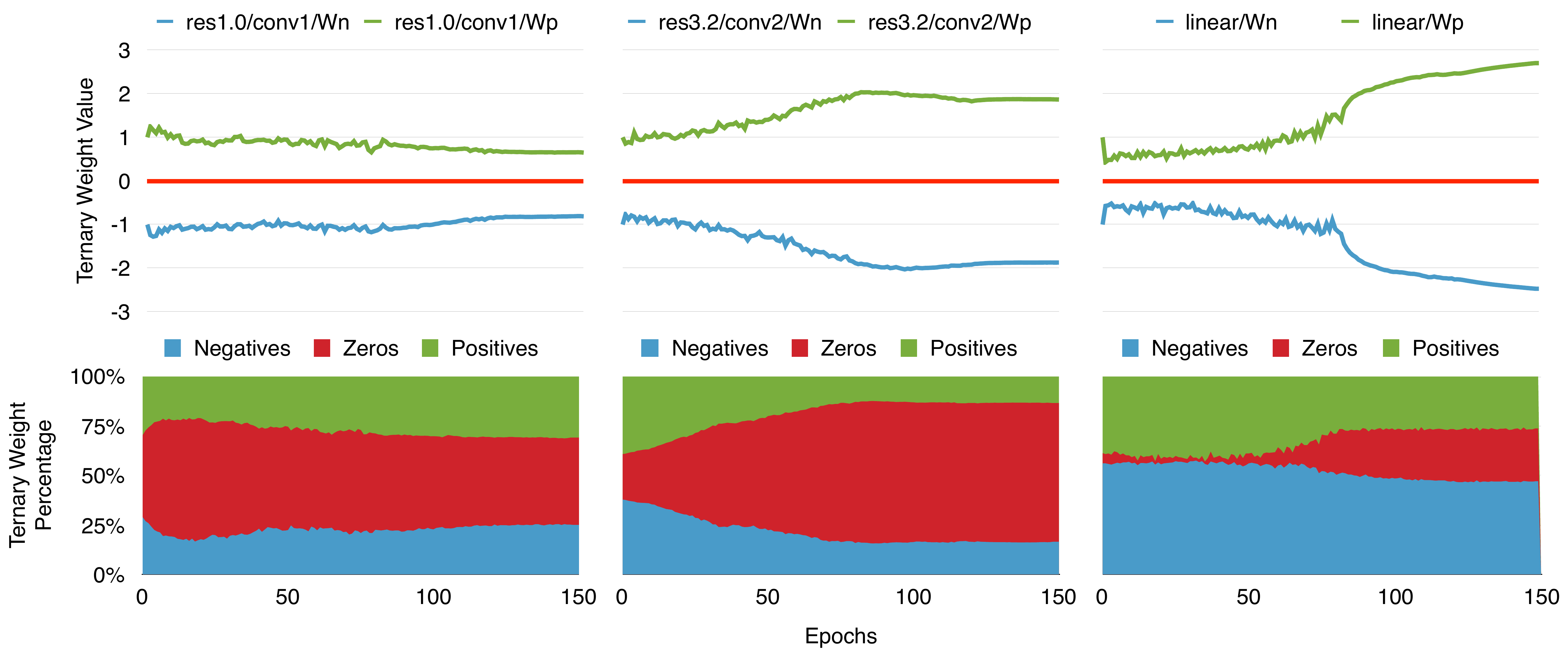}
\caption{Ternary weights value (above) and distribution (below) with iterations for different layers of ResNet-20 on CIFAR-10.}
% \vspace{-10pt}
\label{fig:weights}
% \vspace{-10pt}
\end{figure}

Note we use scalar number 1 as factor of gradients of zero weights. The overall quantization process is illustrated as Figure~\ref{fig:process}. The evolution of the ternary weights from different layers during training is shown in Figure~\ref{fig:weights}. We observe that as training proceeds, different layers behave differently: for the first quantized conv layer, the absolute values of $W^p_l$ and $W^n_l$ get smaller and sparsity gets lower, while for the last conv layer and fully connected layer, the absolute values of $W^p_l$ and $W^n_l$ get larger and sparsity gets higher. 

We learn the ternary assignments~(index to the codebook) by updating the latent full-resolution weights during training. This may cause the assignments to change between iterations. Note that the thresholds are not constants as the maximal absolute values change over time. Once an updated weight crosses the threshold, the ternary assignment is changed. 

% In TWN~\cite{li2016ternary} and other ternary networks, quantized weights in one layer is scaled using same value for positives and negatives. For networks with batch normalization, this uniform scaling has the same effect as the scaling factor in batch normalization, and thus not learning extra information. On the contrary, separate $W^i_p$ and $W^i_n$ scaling factors can help us to scale up or down the positive weights and negative weights separately, which can not be achieved by previous methods. 

% \begin{equation}
% w^q(n,c,w,h)=\left\{
% \begin{aligned}
% W^i_p & : w(n,c,w,h)>w'_{th} \\
% -W^i_n & : w(n,c,w,h)<-w'_{th} \\
% 0 & : otherwise 
% \end{aligned}
% \right.
% \end{equation}

The benefits of using trained quantization factors are: i) The asymmetry of $W^p_l\neq W^n_l$ enables neural networks to have more model capacity. ii) Quantized weights play the role of "learning rate multipliers" during back propagation.

\subsection{Quantization Heuristic}
In previous work on ternary weight networks, \cite{li2016ternary} proposed Ternary Weight Networks (TWN) using $\pm \Delta_l$ as thresholds to reduce 32-bit weights to ternary values, where $\pm \Delta_l$ is defined as Equation~\ref{eq:5}. They optimized value of $\pm \Delta_l$ by minimizing expectation of L2 distance between full precision weights and ternary weights. 
Instead of using a strictly optimized threshold, we adopt different heuristics: 1) use the maximum absolute value of the weights as a reference to the layer's threshold and maintain a constant factor $t$ for all layers:
\begin{equation} \label{eq:t}
\Delta_l=t\times \text{max}(|\tilde{w}|)
\end{equation}
and 2) maintain a constant sparsity $r$ for all layers throughout training. By adjusting the hyper-parameter $r$ we are able to obtain ternary weight networks with various sparsities. We use the first method and set $t$ to 0.05 in experiments on CIFAR-10 and ImageNet dataset and use the second one to explore a wider range of sparsities in section 5.1.1. 

We perform our experiments on CIFAR-10~\citep{krizhevsky2009learning} and ImageNet~\citep{ILSVRC15}. 
Our network is implemented on both TensorFlow~\citep{tensorflow2015-whitepaper} and Caffe~\citep{jia2014caffe} frameworks.

\subsection{CIFAR-10}

\section{Experiments}
\begin{figure}[b]
\vspace{-20pt}
\hspace{-10pt}
\centering
\includegraphics[width=0.9\textwidth]{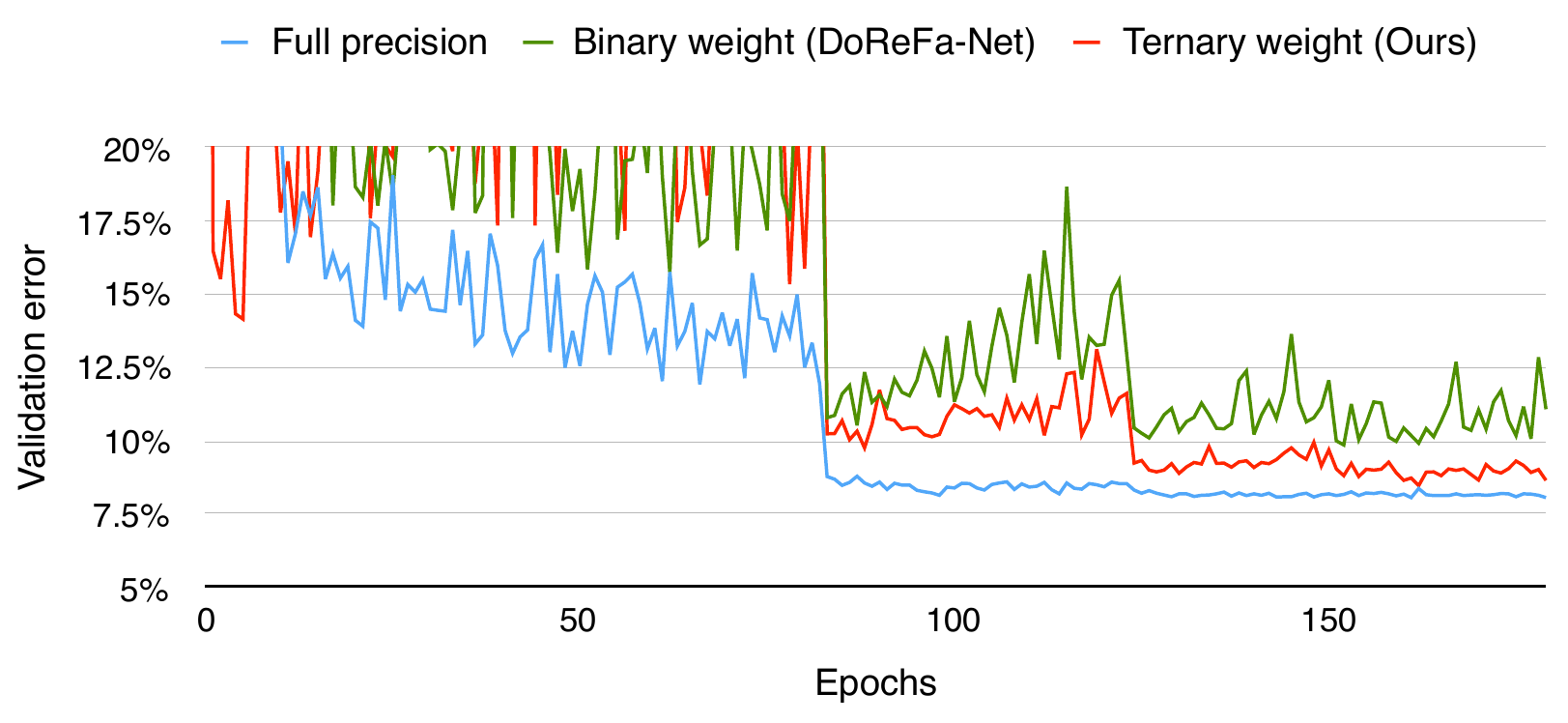}
\caption{ResNet-20 on CIFAR-10 with different weight precision.}
\label{fig:accuracy}
% \vspace{-10pt}
\end{figure}

CIFAR-10 is an image classification benchmark containing images of size 32$\times$32RGB pixels in a training set of 50000 and a test set of 10000. ResNet~\citep{he2015deep} structure is used for our experiments. 

We use parameters pre-trained from a full precision ResNet to initialize our model. Learning rate is set to 0.1 at beginning and scaled by 0.1 at epoch 80, 120 and  300. A L2-normalized weight decay of 0.0002 is used as regularizer. Most of our models converge after 160 epochs. We take a moving average on errors of all epochs to filter off fluctuations when reporting error rate. 

We compare our model with the full-precision model and a binary-weight model. We train a a full precision ResNet~\citep{he2016identity} on CIFAR-10 as the baseline (blue line in Figure~\ref{fig:accuracy}). We fine-tune the trained baseline network as a 1-32-32 DoReFa-Net where weights are 1 bit and both activations and gradients are 32 bits  giving a significant loss of accuracy (green line) . Finally, we fine-tuning the baseline with trained ternary weights (red line). Our model has substantial accuracy improvement over the binary weight model, and our loss of accuracy over the full precision model is small.  We also compare our model to Tenary Weight Network~(TWN) on ResNet-20. Result shows our model improves the accuracy by $\sim 0.25\%$ on CIFAR-10.

We expand our experiments to ternarize ResNet with 32, 44 and 56 layers. All ternary models are fine-tuned from full precision models. Our results show that we improve the accuracy of ResNet-32, ResNet-44 and ResNet-56 by 0.04\%, 0.16\% and 0.36\% . The deeper the model, the larger the improvement. We conjecture that this is due to ternary weights providing the right model capacity and preventing overfitting for deeper networks.

\begin{table}[h]
% \vspace{-10pt}
\centering
\label{DS1Results}
\begin{tabular}{c|c|c|c}
\hline
Model & Full resolution & Ternary (Ours) &  Improvement \\ \hline
ResNet-20 & 8.23 & \bf{8.87} & \bf{-0.64}  \\
ResNet-32 & 7.67 & \bf{7.63} & \bf{0.04} \\
ResNet-44 & 7.18 & \bf{7.02} & \bf{0.16} \\
ResNet-56 & 6.80 & \bf{6.44} & \bf{0.36} \\
% ResNet-110 & 5.97 & \bf{6.23}  \\
\hline
\end{tabular}
% \vspace{-5pt}
\caption{Error rates of full-precision and ternary ResNets on Cifar-10}
\vspace{5pt}
\end{table}

\subsection{ImageNet}
We further train and evaluate our model on ILSVRC12(\cite{ILSVRC15}). ILSVRC12 is a 1000-category dataset with over 1.2 million images in training set and 50 thousand images in validation set. Images from ILSVRC12 also have various resolutions. We used a variant of AlexNet(\cite{NIPS2012_4824}) structure by removing dropout layers and add batch normalization\citep{ioffe2015batch} for all models in our experiments. The same variant is also used in experiments described in the paper of DoReFa-Net.

Our ternary model of AlexNet uses full precision weights for the first convolution layer and the last fully-connected layer. Other layer parameters are all quantized to ternary values. We train our model on ImageNet from scratch using an Adam optimizer (\cite{kingma2014adam}). Minibatch size is set to 128. Learning rate starts at $10^{-4}$ and is scaled by 0.2 at epoch 56 and 64. A L2-normalized weight decay of $5 \times 10^{-6}$ is used as a regularizer. Images are first resized to $256\times 256$ then randomly cropped to $224\times 224$ before input. We report both top 1 and top 5 error rate on validation set.

We compare our model to a full precision baseline, 1-32-32 DoReFa-Net and TWN. After around 64 epochs, validation error of our model dropped significantly compared to other low-bit networks as well as the full precision baseline. Finally our model reaches top 1 error rate of 42.5\%, while DoReFa-Net gets 46.1\% and TWN gets 45.5\%. Furthermore, our model still outperforms full precision AlexNet (the batch normalization version, 44.1\% according to paper of DoReFa-Net) by 1.6\%, and is even better than the best AlexNet results reported (42.8\%\footnote{https://github.com/BVLC/caffe/wiki/Models-accuracy-on-ImageNet-2012-val}). The complete results are listed in Table~\ref{tb:imagenet_data}.

% \begin{table}[h]
% \centering
% \begin{tabular}{c|c|c|c|c|c}
% \hline
% Error & \begin{tabular}[c]{@{}c@{}}Full precision\\ (with Dropout)\end{tabular} & \begin{tabular}[c]{@{}c@{}}Full precision\\ (with BN)\end{tabular} & \begin{tabular}[c]{@{}c@{}}1-bit \\ (DoReFa-Net)\end{tabular} & \begin{tabular}[c]{@{}c@{}}2-bit\\ (TWN)\end{tabular} & \begin{tabular}[c]{@{}c@{}}2-bit\\ (Ours)\end{tabular} \\ \hline
% Top1  & 42.8\%                                                                  & 44.1\%                                                             & 46.1\%                                                        & 45.5\%                                                & \textbf{42.5\%}                                        \\
% Top5  & 19.7\%                                                                  & Not reported                                                       & 23.7\%                                                        & 23.2\%                                                & \textbf{20.3\%}   \\           \hline                        
% \end{tabular}

% \caption{Top1 and Top5 error rate of AlexNet on ImageNet}
% \label{tb:imagenet_data}
% \end{table}

\begin{table}[h]
\centering
\begin{tabular}{c|c|c|c|c}
\hline
Error & \begin{tabular}[c]{@{}c@{}}Full precision \end{tabular} &  \begin{tabular}[c]{@{}c@{}}1-bit \\ (DoReFa)\end{tabular} & \begin{tabular}[c]{@{}c@{}}2-bit\\ (TWN)\end{tabular} & \begin{tabular}[c]{@{}c@{}}2-bit\\ (Ours)\end{tabular} \\ \hline
Top1  & 42.8\% & 46.1\%  & 45.5\% & \textbf{42.5\%}                                        \\
Top5  & 19.7\% & 23.7\%  & 23.2\% & \textbf{20.3\%}   \\           \hline                        
\end{tabular}

\caption{Top1 and Top5 error rate of AlexNet on ImageNet}
\label{tb:imagenet_data}
\end{table}

\begin{figure}[t]
\vspace{-20pt}
\hspace{0pt}
\centering
\includegraphics[width=1.00\textwidth]{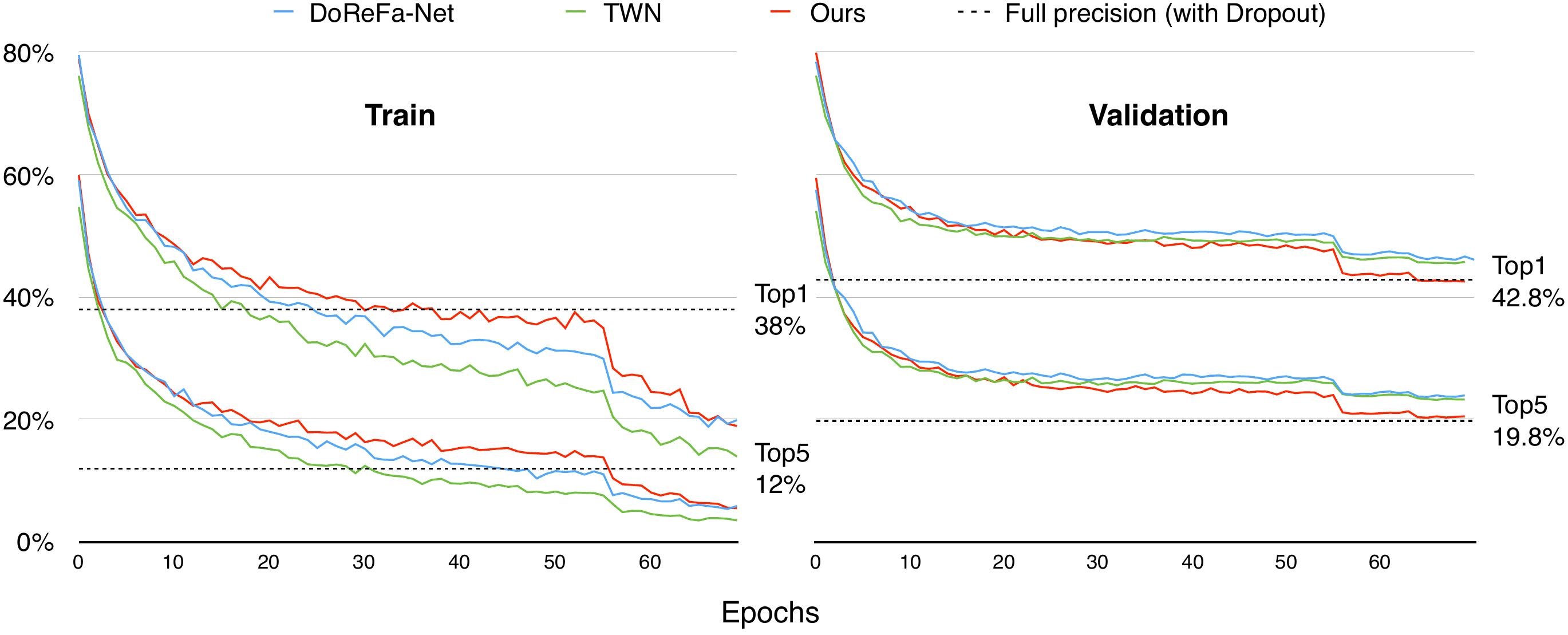}
\vspace{-10pt}
\caption{Training and validation accuracy of AlexNet on ImageNet}
\label{fig:imagenet_accuracy}
% \vspace{-20pt}
\end{figure}

We draw the process of training in Figure \ref{fig:imagenet_accuracy}, the baseline results of AlexNet are marked with dashed lines. Our ternary model effectively reduces the gap between training and validation performance, which appears to be quite great for DoReFa-Net and TWN. This indicates that adopting trainable $W^p_l$ and $W^n_l$ helps prevent models from overfitting to the training set.

We also report the results of our methods on ResNet-18B in Table~\ref{tb:imagenet_resnet}. The full-precision error rates are obtained from Facebook's implementation. Here we cite Binarized Weight Network(BWN)\cite{rastegari2016xnor} results with all layers quantized and TWN finetuned based on a full precision network, while we train our TTQ model from scratch. Compared with BWN and TWN, our method obtains a substantial improvement. 

\begin{table}[h]
\centering
\begin{tabular}{c|c|c|c|c}
\hline
Error & \begin{tabular}[c]{@{}c@{}}Full precision \end{tabular} &  \begin{tabular}[c]{@{}c@{}}1-bit \\ (BWN)\end{tabular} & \begin{tabular}[c]{@{}c@{}}2-bit\\ (TWN)\end{tabular} & \begin{tabular}[c]{@{}c@{}}2-bit\\ (Ours)\end{tabular} \\ \hline

Top1  & 30.4\% & 39.2\%  & 34.7\% & \textbf{33.4\%}                                        \\
Top5  & 10.8\% & 17.0\%  & 13.8\% & \textbf{12.8\%}   \\           \hline                        
\end{tabular}

\caption{Top1 and Top5 error rate of ResNet-18 on ImageNet}
\label{tb:imagenet_resnet}
\end{table}

\section{Discussion}
In this section we analyze performance of our model with regard to weight compression and inference speeding up. These two goals are achieved through reducing bit precision and introducing sparsity. We also visualize convolution kernels in quantized convolution layers to find that basic patterns of edge/corner detectors are also well learned from scratch even precision is low.

\subsection{Spatial and energy efficiency}
We save storage for models by $16\times$ by using ternary weights. Although switching from a binary-weight network to a ternary-weight network increases bits per weight, it brings sparsity to the weights, which gives potential to skip the computation on zero weights and achieve higher energy efficiency.

\begin{figure}[t]
\vspace{-20pt}
\hspace{0pt}
\centering
\includegraphics[width=0.85\textwidth]{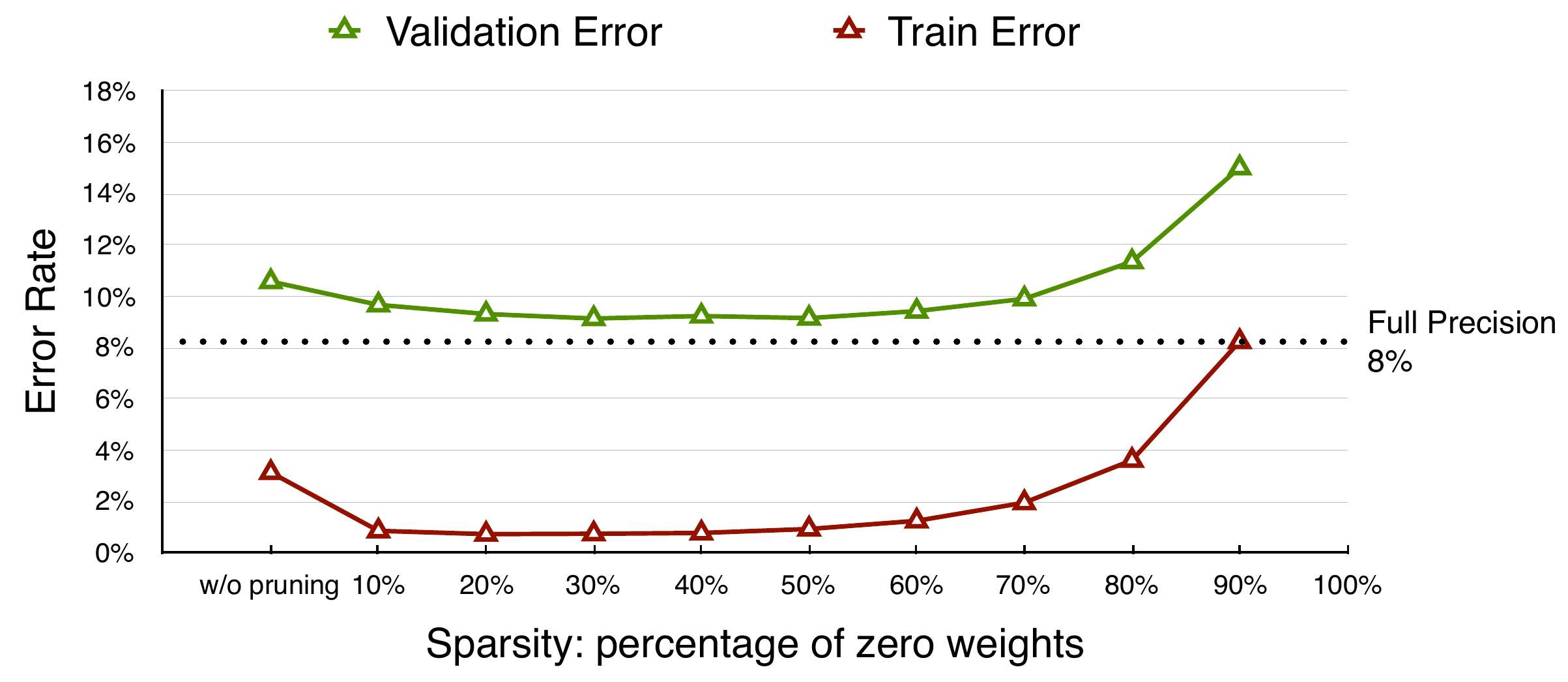}
\vspace{-10pt}
\caption{Accuracy v.s. Sparsity on ResNet-20}
\label{fig:sparsity}
\vspace{-20pt}
\end{figure}

% \usepackage{multirow}
\begin{table}[b]
\centering
\begin{tabular}{|c|cc|cc|cc|}
\multirow{2}{*}{Layer} & \multicolumn{2}{c|}{Full precision} & \multicolumn{2}{c|}{Pruning (NIPS'15)} & \multicolumn{2}{c|}{Ours} \\
                       & Density     & Width      & Density       & Width       & Density  & Width   \\ \hline
conv1                  & 100\%       & 32 bit     & 84\%          & 8 bit      & 100\%    & 32 bit  \\
conv2                  & 100\%       & 32 bit     & 38\%          & 8 bit      & 23\%     & 2 bit   \\
conv3                  & 100\%       & 32 bit     & 35\%          & 8 bit      & 24\%     & 2 bit   \\
conv4                  & 100\%       & 32 bit     & 37\%          & 8 bit      & 40\%     & 2 bit   \\
conv5                  & 100\%       & 32 bit     & 37\%          & 8 bit      & 43\%     & 2 bit   \\ \hline
conv total             & 100\%       & -           & 37\%          & -          & 33\%     & -       \\ \hline
fc1                    & 100\%       & 32 bit     & 9\%           & 5 bit      & 30\%     & 2 bit   \\
fc2                    & 100\%       & 32 bit     & 9\%           & 5 bit      & 36\%     & 2 bit   \\
fc3                    & 100\%       & 32 bit     & 25\%          & 5 bit      & 100\%    & 32 bit  \\ \hline
fc total               & 100\%       & -          & 10\%          & -          & 37\%     & -        \\ \hline
All total              & 100\%       & -          & 11\%          & -          & 37\%     & -       
\end{tabular}
\caption{Alexnet layer-wise sparsity}
\label{sparsity_alexnet}
\end{table}

\subsubsection{Trade-off between sparsity and accuracy}
%\cite{han2015learning} proposed using pruning to zero out small weights by a fixed percentage from full precision networks. Following this idea, we experimented the relationship between error rate and sparsity. 
Figure~\ref{fig:sparsity} shows the relationship between sparsity and accuracy. As the sparsity of weights grows from 0~(a pure binary-weight network) to 0.5~(a ternary network with 50\% zeros), both the training and validation error decrease. Increasing sparsity beyond 50\% reduces the model capacity too far, increasing error.  Minimum error occurs with sparsity between 30\% and 50\%. 
\begin{comment}
As pointed out by Han~\cite{han2015deep}, sparsity contributes to both model compression and computing speed. The relationship between sparsity and accuracy enables people to trade-off accuracy for speed and space or the other way around.
\end{comment}

We introduce only one hyper-parameter to reduce search space. This hyper-parameter can be either sparsity, or the threshold $t$ w.r.t the max value in Equation~6.
%~\ref{eqn:weight-assign}. 
We find that using threshold produces better results. This is because fixing the threshold allows the sparsity of each layer to vary (Figure~ref{fig:weights}). 

\subsubsection{Sparsity and efficiency of AlexNet}

We further analyze parameters from our AlexNet model. We calculate layer-wise density (complement of sparsity) as shown in Table \ref{sparsity_alexnet}.  Despite we use different $W^p_l$ and $W^n_l$ for each layer, ternary weights can be pre-computed when fetched from memory, thus multiplications during convolution and inner product process are still saved. Compared to Deep Compression, we accelerate inference speed using ternary values and more importantly, we reduce energy consumption of inference by saving memory references and multiplications, while achieving higher accuracy.

We notice that without all quantized layers sharing the same $t$ for Equation \ref{eq:t}, our model achieves considerable sparsity in convolution layers where the majority of computations takes place. Therefore we are able to squeeze forward time to less than 30\% of full precision networks. 

As for spatial compression, by substituting 32-bit weights with 2-bit ternary weights, our model is approximately $16\times$ smaller than original 32-bit AlexNet.

\subsection{Kernel Visualization}
\begin{figure}[t]
\hspace{0pt}
\centering
\includegraphics[width=1.00\textwidth]{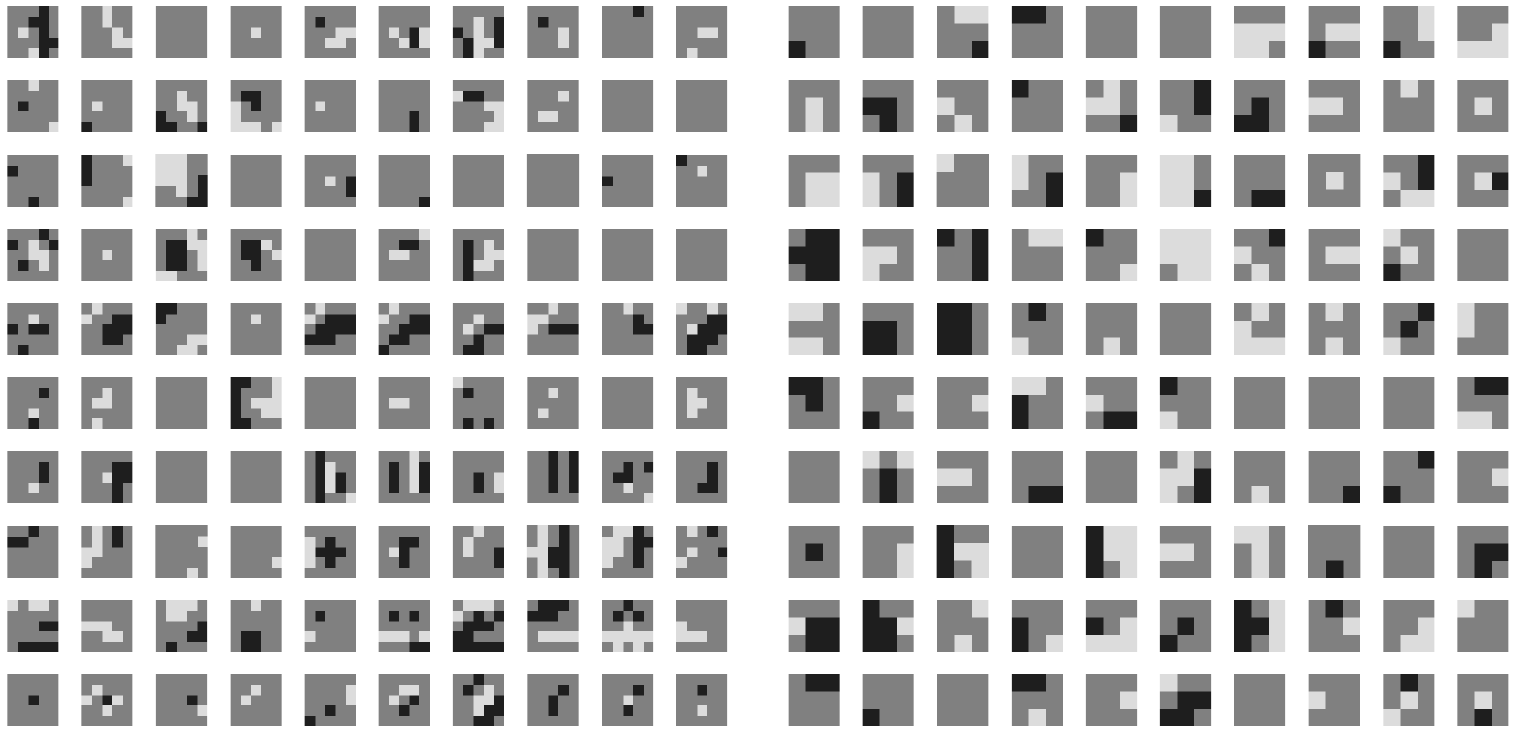}
\vspace{-10pt}
\caption{Visualization of kernels from Ternary AlexNet trained from Imagenet.}
\label{fig:kernels}
% \vspace{-20pt}
\end{figure}

We visualize quantized convolution kernels in Figure \ref{fig:kernels}. The left matrix is kernels from the second convolution layer ($5\times 5$) and the right one is from the third ($3\times 3$). We pick first 10 input channels and first 10 output channels to display for each layer. Grey, black and white color represent zero, negative and positive weights respectively. 

We observe similar filter patterns as full precision AlexNet. Edge and corner detectors of various directions can be found among listed kernels. While these patterns are important for convolution neural networks, the precision of each weight is not. Ternary value filters are capable enough extracting key features after a full precision first convolution layer while saving unnecessary storage.

Furthermore, we find that there are a number of empty filters (all zeros) or filters with single non-zero value in convolution layers. More aggressive pruning can be applied to prune away these redundant kernels to further compress and speed up our model.

\section{Conclusion}

We introduce a novel neural network quantization method that compresses network weights to ternary values. We introduce two trained scaling coefficients $W^l_p$ and $W^l_n$ for each layer and train these coefficients using back-propagation.   During training, the gradients are back-propagated both to the latent full-resolution weights and to the scaling coefficients. We use layer-wise thresholds that are proportional to the maximum absolute values to quantize the weights. When deploying the ternary network, only the ternary weights and scaling coefficients are needed, which reducing parameter size by at least $16\times$. Experiments show that our model reaches or even surpasses the accuracy of full precision models on both CIFAR-10 and ImageNet dataset.  On  ImageNet we exceed the accuracy of prior ternary networks (TWN) by 3\%.

\newpage
\footnotesize
\bibliography{ref}
\bibliographystyle{iclr2017_conference}

\end{document}